\documentclass{article}
\usepackage[preprint]{neurips_2019}

\usepackage[utf8]{inputenc} % allow utf-8 input
\usepackage[T1]{fontenc}    % use 8-bit T1 fonts
\usepackage{hyperref}       % hyperlinks
\usepackage{url}            % simple URL typesetting
\usepackage{booktabs}       % professional-quality tables
\usepackage{amsfonts}       % blackboard math symbols
\usepackage{nicefrac}       % compact symbols for 1/2, etc.
\usepackage{microtype}      % microtypography

\usepackage{url}
\usepackage{graphicx}
\usepackage{amsmath}
\usepackage{booktabs}
\usepackage{algorithm}
\usepackage{algorithmic}
\usepackage{booktabs}
\usepackage{array}
\usepackage{color}
\usepackage{amsmath,amssymb,amsfonts}
\usepackage{bm}
\usepackage{xcolor}
\usepackage{subfigure}
\usepackage{balance}
\usepackage{multirow}
\usepackage{authblk}

\title{Label Mapping Neural Networks with Response Consolidation for Class Incremental Learning}

\author[1]{\textbf{Xu Zhang}}
\author[1]{\textbf{Yang Yao}}
\author[1]{\textbf{Baile Xu}}
\author[1]{\textbf{Lekun Mao}}
\author[1]{\textbf{Furao Shen}}
\author[2]{\textbf{Jian Zhao}}
\author[3]{\textbf{Qingwei Lin}}

\affil[1]{\footnotesize Department of Computer Science and Technology, Nanjing University, Nanjing, China}
\affil[2]{\footnotesize School of electronic science and engineering, Nanjing University,  Nanjing, China}
\affil[3]{\footnotesize Microsoft Research, Beijing, China}
\affil[ ]{}
\affil[ ]{\texttt{\{zhangxu037,yaoyang,bixu,lekmao\}@smail.nju.edu.cn} }
\affil[ ]{ \texttt { \{frshen,jianzhao\}@nju.edu.cn}} 
\affil[ ]{ \texttt { qlin@microsoft.com}}

\begin{document}

\maketitle

\begin{abstract}
Class incremental learning refers to a special multi-class classification task,  in which the number of classes is not fixed but is increasing with the continual arrival of  new data. Existing researches mainly focused on solving catastrophic forgetting problem in class incremental learning. To this end, however, these models still require the old classes cached in the auxiliary data structure or models, which is inefficient in space or time. In this paper, it is the first time to discuss the difficulty without support of old classes in class incremental learning, which is called as \emph{softmax suppression problem}. To address these challenges, we develop a new model named \emph{Label Mapping with Response Consolidation (LMRC)}, which need not access the old classes anymore. We propose the Label Mapping algorithm combined with the multi-head neural network for mitigating the softmax suppression problem, and propose the Response Consolidation method to overcome the catastrophic forgetting  problem. Experimental results on the benchmark datasets show that our proposed method achieves much better performance compared to the related methods in different scenarios.
\end{abstract}

\section{Introduction}
\label{introduction}

Over the years, the fast development of Deep Neural Network (DNN) has been witnessed in various fields \cite{he2016deep, krizhevsky2012imagenet}.
However, there are still many challenges that hinder developing a DNN-based intelligence system targeted at real-world applications.
One of these challenges is how to enable deep networks to learn incremental classes from streaming data, like the way human beings learn new concepts from daily life.
It is also called as \textbf{class incremental learning}, where batches of labelled data of new classes are made available gradually \cite{xiao2014error}.
We expect the model should be updated continually so that the knowledge embedded in new classes can be incorporated without sacrificing the learned knowledge of old classes too much~\cite{zhou2002hybrid}.

The class incremental learning of neural networks has been studied since a long time ago.
It is widely recognized that the \textbf{catastrophic forgetting} phenomenon \cite{mccloskey1989catastrophic} is the biggest obstacle for neural networks to keep the memory of old classes. 
The reason for this notorious problem is that the crucial network weights for old classes are changed to meet the objectives of new arriving classes \cite{kirkpatrick2017overcoming}.
Catastrophic forgetting also occurs in \textit{task incremental learning}~\cite{van2018generative} and \textit{reinforcement continual learning}~\cite{kirkpatrick2017overcoming},  where the training data of different tasks are also fed incrementally.

A general and straightforward solution to the catastrophic forgetting problem is \emph{Rehearsal}~\cite{robins1998catastrophic}: the neural network is able to retrain by the old classes while learning new classes.
In general, the methods of this category need to utilize a \emph{data pool} to cache the old classes, or a \emph{generative model} to generate some pseudo old training samples for retraining \cite{robins1998catastrophic,shin2017continual,van2018generative}.
Rehearsal can significantly alleviate the catastrophic forgetting problem, but it also needs to retrain the model with the data of the old classes and requires extra storage space. 
However,  the motivation of class incremental learning is aiming to reduce the training load by a progressive learning way~\cite{zhou2002hybrid}. 
In other words, the model should be  better trained based on its current parameters and the data of new classes  instead of the data of all classes.
In addition, in some scenarios, such as data streams, the model may not be able to access the data of some old classes at all. 

For overcoming catastrophic forgetting without the help of Rehearsal,  some excellent methods are recently proposed and achieve outstanding results, including Elastic Weight Consolidation (EWC)~\cite{kirkpatrick2017overcoming}, Learning without Forgetting (LwF) ~\cite{li2017learning}, Synaptic Intelligence (SI)~\cite{zenke2017continual}, etc.
However, almost all of these models are focused on task incremental learning or  reinforcement continual learning.
When applied to class incremental learning problem directly, these models suffer from another intractable problem. We call it as softmax suppression, which cannot be solved merely by overcoming catastrophic forgetting.

Softmax suppression refers that the output probabilities of the new classes suppress that of the old classes in the classification layer.
The main reason for this phenomenon is : a) The network is trained without the support of Rehearsal, i.e. the network being trained cannot access the old classes anymore. b) New classes and old classes are trained in the same softmax layer. 

\begin{figure}[htbp]
\vspace{-18pt}
\centering
\begin{minipage}[t]{0.48\textwidth}
\centering
\includegraphics[height=0.65\textwidth,width=\textwidth]{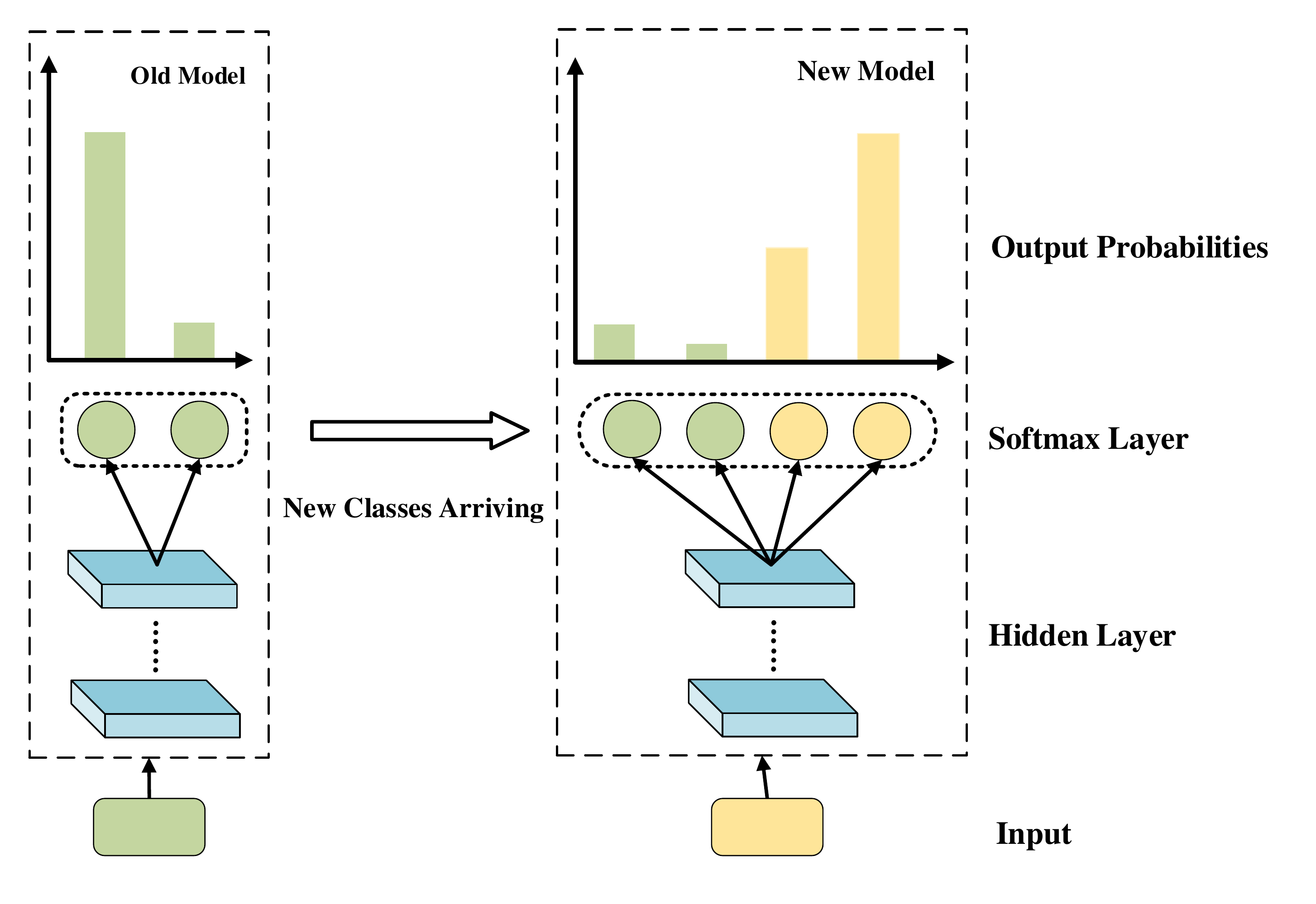}
\end{minipage}
\begin{minipage}[t]{0.48\textwidth}
\centering
\includegraphics[height=0.7\textwidth,width=1\textwidth]{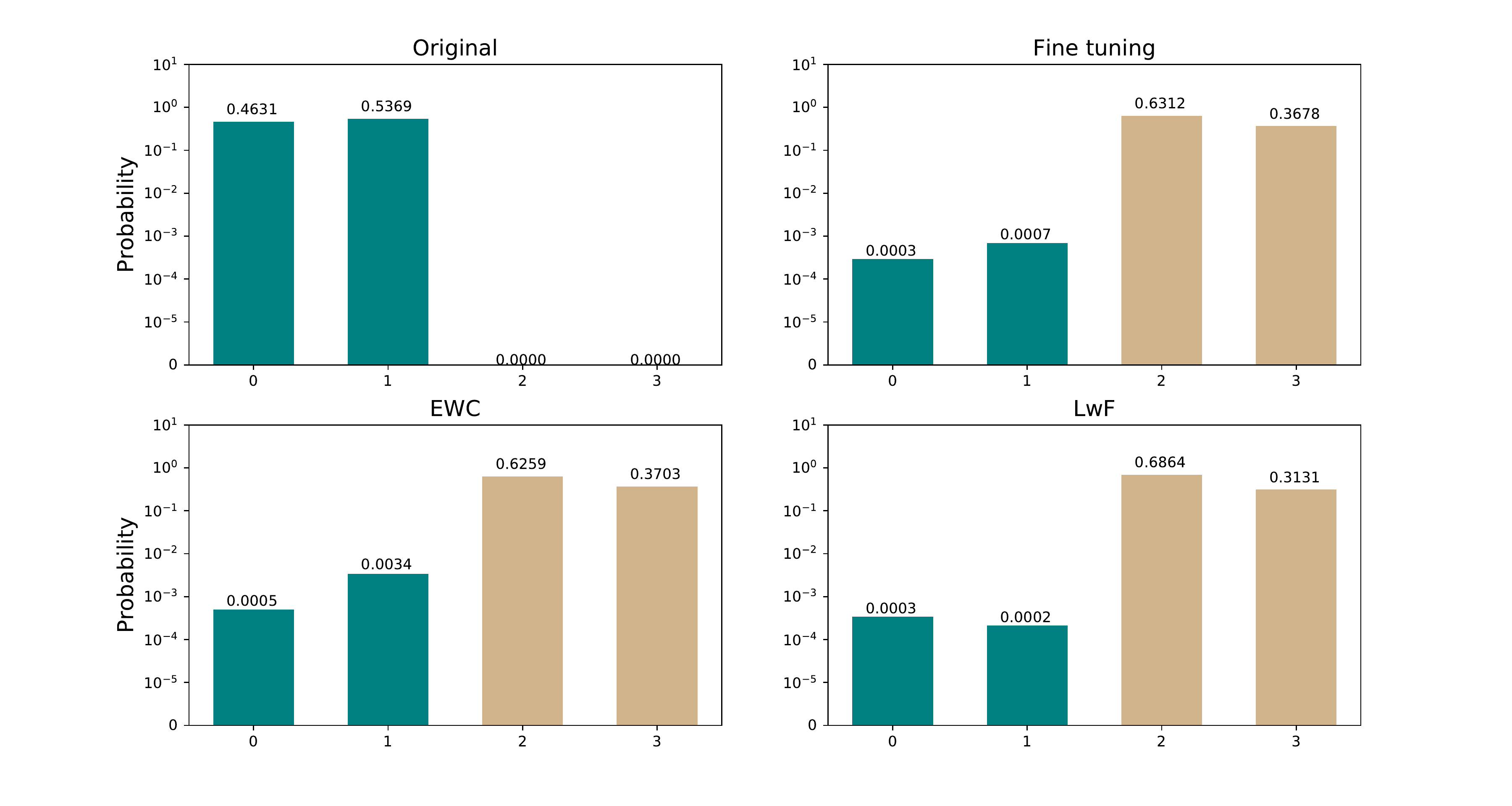}
\label{softmax_problem_explore}
\end{minipage}
\caption{\textbf{Left}: Example of softmax suppression. The green neural units represent old classes and the yellow neural units represent new classes. \textbf{Right}: Softmax Suppression Problem Exploration Result.}
\label{softmax}
\vspace{-13pt}
\end{figure}

We illustrate an example of softmax suppression  in Fig.~\ref{softmax} (Left).
Owing to the first point, the output probabilities of new classes given by the network definitely exceed that of old classes. It is because that the network should output the higher probabilities of new classes for meeting the training objectives. Due to the second point, the higher probabilities of new classes would inevitably lead to lower probabilities of old classes because the sum of all probabilities is 1. 
Therefore, as long as the model is unable to access the data of old classes during training, whatever methods we adopt for preventing the catastrophic forgetting, this suppression effect would still clamp the output probabilities of the old classes so that we cannot predict them.

We conduct an exploration experiment on the MNIST dataset to verify this problem. 
We firstly train a CNN model on the first two classes. Then we expand the output dimension and keep training the network on the latter two classes.
During the training process of the latter two classes,  the model cannot access the previous two classes anymore.
Except simple fine tuning on the final classification layer, we also try to utilize EWC and LwF method to prevent the network from forgetting the old classes.
After training, we test the model on all four  classes, including the old two classes and new two classes. 
The average output probabilities on four positions of final softmax layer are shown in  Fig.~\ref{softmax} (Right).
It is obvious that whatever anti-forgetting techniques we adopt, after training on the latter two classes, the  output probabilities tend to concentrate on the new classes. 
Even if we fed the testing samples of the old classes into the network, the model still recognizes them as the new classes.

In summary,  in order to achieve class incremental learning without Rehearsal, a new model is required to handle catastrophic forgetting and softmax suppression simultaneously.
In this paper, we propose a new method, named \textbf{Label Mapping with Response Consolidation (LMRC)}.
We propose\emph{ Label Mapping (LM)} algorithm to mitigate the softmax suppression problem and \emph{Response Consolidation (RC) }algorithm to handle the catastrophic forgetting problem.
The highlight of this paper is that we achieve class incremental learning of deep neural networks \textbf{without accessing the data of old classes anymore}.
Moreover, if allowed, the accuracy of LMRC could also be further improved with Rehearsal without any changes to the model.

\section{Proposed Approach}
\label{approach}
\subsection{Overview}
\begin{figure}[htbp]
\centering
%\subfloat[]{
\begin{minipage}[t]{0.48\textwidth}
\centering
\includegraphics[height=0.65\textwidth,width=1\textwidth]{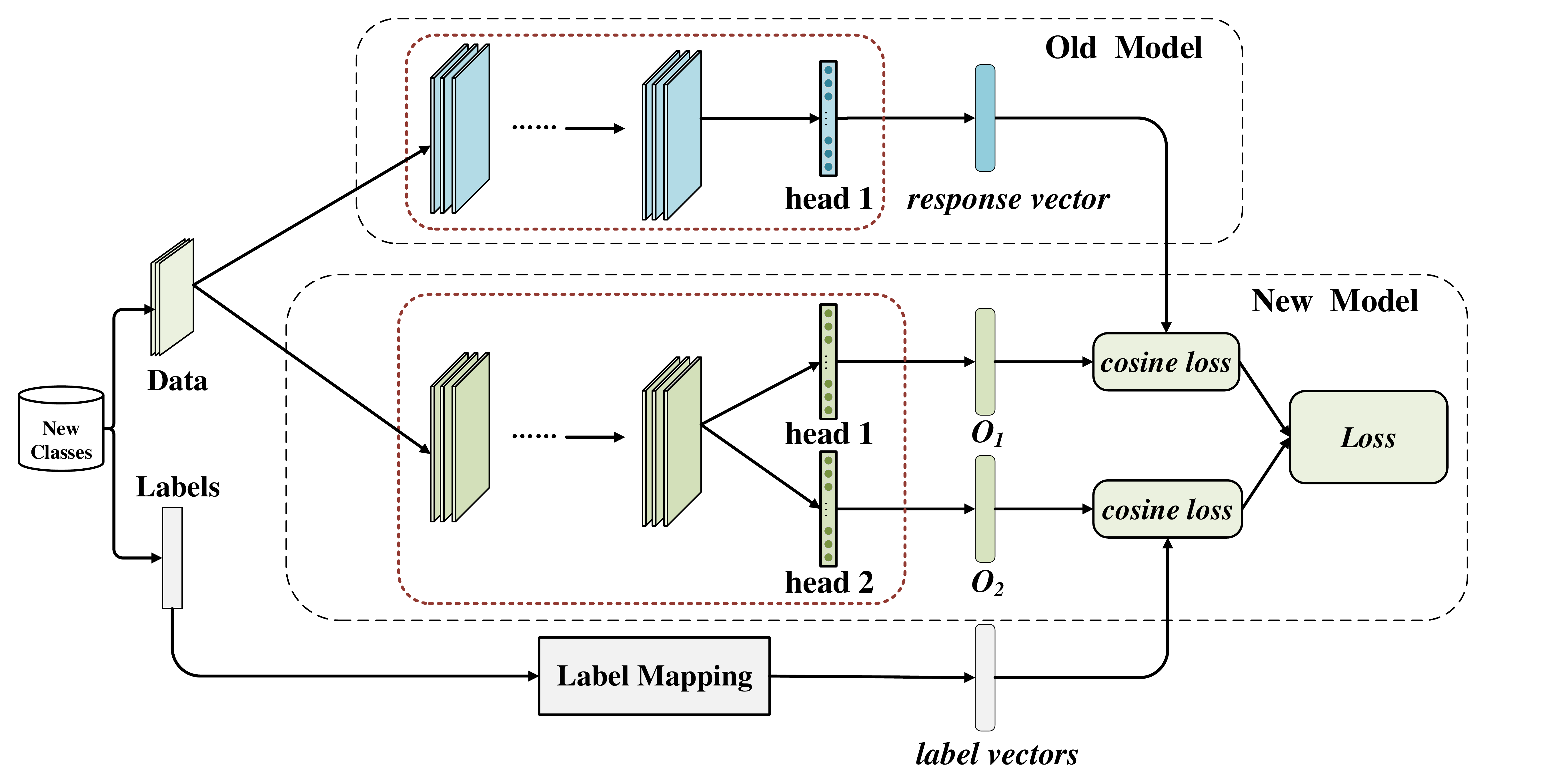}
%\label{LMRC}
\end{minipage}
%}
%\subfloat[]{
\begin{minipage}[t]{0.48\textwidth}
\centering
\includegraphics[height=0.6\textwidth,width=1.1\textwidth]{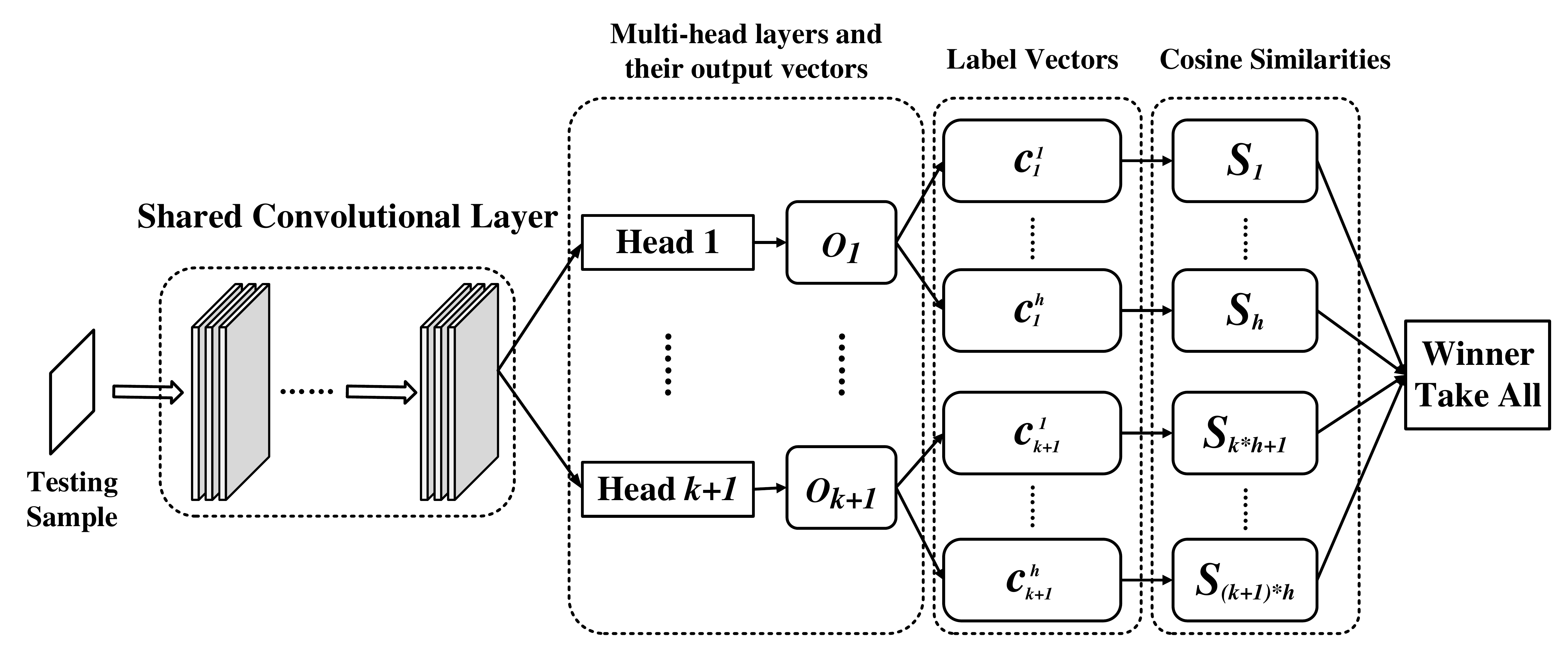}
%\label{predict}
\end{minipage}
%}
\caption{\textbf{Left}: The framework of LMRC. \textbf{Right}: The testing process of LMRC. $\bm{o}$ represents the output vector from a head, $\bm{c}_{i}^{j}$ represents the label vector, $\bm{S}$ represents the  cosine similarity.}
\label{LMRC}
\vspace{-15pt}
\end{figure}

In this paper, we propose a novel and universal class incremental learning model, named \emph{Label Mapping with Response Consolidation} \emph{(LMRC)}.
The framework of  LMRC is shown in Fig.~\ref{LMRC} (Left).
LMRC utilizes a multi-head neural network as the basic architecture, as shown in the red boxes of Fig.~\ref{LMRC} (Left).
The \emph{Old Model} (blue part) represents the previous network trained on the old classes,  and the \emph{New Model} (green part) represents the network being trained on the new classes.
When new classes  arrive, a new head is accordingly added to the \emph{New Model} for training these new classes, like \emph{head 2} shown in Fig.~\ref{LMRC} (Left).
After that,  the  training data of the new classes are propagated through all old heads (i.e. \emph{head 1} ) of the \emph{Old Model} and the outputs are denoted as the \emph{response vectors}.
Then, a set of \emph{label vectors} are generated by the Label Mapping algorithm (gray part), which can be regarded as the labels of the new classes data.
The label vectors and the response vectors are the desired targets of the new head and the old heads in the \emph{New Model}, respectively.
Therefore,  we train the network to make the outputs of these heads ($O_1$ and $O_2$ in Fig.~\ref{LMRC}  (Left)) to approximate these targets, respectively.

% \begin{figure}[h!]
% \centering
% \subfigure[]{
% \includegraphics[height=0.35\textwidth,width=0.48\textwidth]{image/LMRC2.pdf}
% }
% \subfigure[]{
% \includegraphics[height=0.35\textwidth,width=0.48\textwidth]{image/LMRC_predict.pdf}
% }
% \caption{Classification accuracy of LMRC and LM}
% \label{RC}
% \end{figure}

% \begin{figure*}[h!]
% \vspace{0pt}
%   \centering
%   \includegraphics[height=0.4\textwidth,width=.9\textwidth]{image/LMRC2.pdf}
%   \caption{The framework of LMRC. }
%   \label{LMRC}
% \end{figure*}

\subsection{Multi-head network structure}
\label{CT2MT}
The softmax suppression problem described in Sec. \ref{introduction} is a great obstacle in class incremental learning. 
The reason for this problem is that the old classes and the new classes share the same classification layer during training and testing.
In order to deal with this problem, we seek inspirations from the multi-task deep learning model, which owns multiple \textit{heads} on top of the network. 
Each head is a single softmax layer for classification.
Using the multi-head network, the new classes and the old classes can be assigned to different heads for training and predicting.  
For example, when we encounter new classes in the streaming data, we open up a new head on the top of the network to train these new classes, like \textit{head 2} in Fig.~\ref{LMRC} (Left). 
We can say that this head \emph{governs} these new classes. 
In this way, the probabilities of new classes do not suppress those of old classes anymore because they are not trained in the same softmax layer.

Of course, this solution faces an important problem. 
Different from the multi-task learning, in the testing phase, we do not know which head should be used to predict the input samples. 
For example,  \textbf{Head A} governs the \textbf{Class 1} and \textbf{Head B} governs  the \textbf{Class 2}.  
If a testing sample is fed into the multi-head network and given the same high probability by both of heads, the model would not be able to judge which one should be chosen as the final output. 
We call it as \textbf{Confusion Problem.}
As the number of heads increasing, or the number of classes governed by the heads increasing,  Confusion Problem would become more and more significant.

The reason behind this problem is that the traditional softmax layer suffers from high \textbf{intra-class variations}~\cite{wen2016discriminative}.
Each head cannot recognize the unknown classes that not governed by it.
It means that each head will unexpectedly output  high probabilities of the classes that are not governed by it. 
In order to solve this problem, we propose the Label Mapping algorithm to reduce the intra-class variations and improve the distinctiveness among the heads.

\subsection{Label Mapping}
\subsubsection{Label Mapping Algorithm}
To enhance the intra-class compactness, we replace the tradition softmax layer by the combination of label vectors and negative cosine similarity loss.
In general classification task, we always denote classes by one-hot coding.
The one-hot codes of different classes are mutually orthogonal.
In LMRC, we still denote the classes by high dimensional vectors, i.e. \emph{label vectors}.
Nevertheless, these label vectors do not have to be strictly orthogonal to each other, but they should still guarantee enough distinctiveness to distinguish different classes.
The pseudo code of Label Mapping is shown in Algorithm \ref{alg1}.

\begin{algorithm}
    \caption{Label Mapping}
    \label{alg1}
        \begin{algorithmic}[1]
            \REQUIRE ~~\\
            New Classes Number, $N$;
            Label Vector Dimension, $d$;
            Similarity Threshold, $T$;
            Max sampling times, $\gamma$; 
            Label Candidate Vectors Set, $S$;
            \ENSURE ~~\\
            Label Candidate Vectors Set, $S$;
			\STATE {$N_{cur} = S.size()$;}
            \STATE $try\_time$ = 0; \\
            \WHILE{$S.size() - N_{cur} < N \wedge  try\_time \leqslant\gamma$}
            	
                  \STATE Sample a vector $\bm{v_{s}} \in \mathbb{R}^{d}$  randomly; \label{marker}
                  \IF{$S = \emptyset $}
                      \STATE $S = S \cup \bm{v_{s}}$
                  \ELSIF{$\frac{ \bm{v_{s}} \cdot \bm{v_{c}}  }{ \left \| \bm{v_{s}} \right \| \   \left \| \bm{v_{c}}\right \| } \leqslant  T, \forall \bm{v_{c}} \in S $}
                      \STATE $S = S \cup \bm{v_{s}}$;\\
                      \STATE $try\_time = 0$; \\
                  \ELSE
                  	  \STATE $try\_time ++$;  \\
%                       \STATE \textbf{goto} Step \emph{\ref{marker}};
                  \ENDIF                
            \ENDWHILE
           \RETURN {$S$;}
    \end{algorithmic}
\end{algorithm}

The main idea of Label Mapping is to generate a set of random vectors for classes denotation.
In order to satisfy the  distinctiveness requirement, we specify a threshold $T$, which controls the upper bound of the cosine similarities among the label vectors.
We randomly sample a unit vector $\bm{v_{s}} \in \mathbb{R}_{d}$ under uniform distribution~\cite{muller1959note}.
If the candidate set $S$ is empty, $\bm{v_{s}}$ will be added into it directly.
Otherwise, $\bm{v_{s}}$ needs to be compared with all the vectors in $S$.
If the cosine similarity between any candidate vector in $S$ and $\bm{v_{s}}$ is higher than $T$, $\bm{v_{s}}$ should be discarded.
Then a new vector $\bm{v_{s}}$ is randomly re-sampled and the above process is repeated.
This process goes on until enough label vectors are generated, or the $try\_time$ has reached the upper limitation $\gamma$.
Finally, each label vector in $S$ is assigned to a specific class.

\subsubsection{Training Phase with Label Vectors}

During the training phase, we use label vectors instead of one-hot codes as the learning targets of our neural network.
Each head in our multi-head network is a normalized linear layer, thus the output of each head is a normalized $d$-dimensional vector instead of probabilities.
We use the \emph{negative cosine similarity} between the output vector and the ground truth, i.e. label vector, as the loss function for new classes.

To be specific, we denote $\mathbf{M}$ as the multi-head neural network, and $\mathbf{M_{i}} (i=1,...,k)$ as the $i\text{-}$th existing head of the network.
Without loss of generality, we assume that each head governs $h$ classes.
When a new classes dataset  $\bm{X}$, which also contains $h$ classes, emerges in training set, a new head $\mathbf{M_{k+1}}$ is added to the network accordingly.
 Meanwhile, a set of label vectors are generated according to Algorithm \ref{alg1} for these new classes.
 We denote $\bm{c}_{i}^{j}$ as the label vector of the $j$-th classes governed by the $i$-th head.
We expect that $\mathbf{M_{k+1}}$ can correctly classify the new classes, therefore we force the output vector $\bm{o}_{k+1}^j$ of  sample $ \bm{x}^j $  ($j$  represents $j$-th class) to move \textbf{closer} to the corresponding label vectors $\bm{c}_{k+1}^{j}$, while keeping them away from the label vectors of old classes in other old heads.
By this way, we compel the output vectors of the network to concentrate on the neighborhood of corresponding label vectors for enhancing the \textbf{intra-class compactness}.
Meanwhile, by the advantage of the Label Mapping, the label vectors are distinguishable enough so that the \textbf{inter-class difference} can also be guaranteed.
The loss function of $\mathbf{M_{k+1}}$ can be defined as Eq~\eqref{equ:1}.
We call it as \textbf{New Classes Loss}.
\begin{equation}
\label{equ:1}
\begin{aligned}
\bm{o}_{k+1}^j &\leftarrow  \mathbf{M_{k+1}}\left ( \bm{x}^j \right ) \\
L_{k+1} &= \frac{  \bm{o}_{k+1}^j \cdot \bm{c}_{k+1}^{j} }{ \left \| \bm{o}_{k+1}^j \right \| \     \left \|  \bm{c}_{k+1}^{j}\right \| }.
\end{aligned}
\end{equation}

\subsubsection{Testing Phase with Label Vectors}

Since the Confusion Problem is mitigated by Label Mapping, we can adopt the \textbf{Winner-Take-All (WTA)} strategy in the testing phase, as shown in Fig.~\ref{LMRC} (Right).
The testing samples are fed into the network and forward propagated through \textbf{all heads} (Head 1 to Head $k+1$ in Fig.~\ref{LMRC} (Right)).
We calculate the similarity between the output vector of each head ($ \bm{o}_{i} $) and the label vectors governed by the corresponding heads ($\bm{c}_{i}^{j}$), so that we can obtain $(k+1) \ast h$ similarity values in total, as shown in Eq~\eqref{sim}.
\begin{equation}
\label{sim}
S_{t} = \frac{  \bm{o}_{i} \cdot \bm{c}_{i}^{j} }{ \left \| \bm{o}_{i} \right \| \     \left \| \bm{c}_{i}^{j}\right \| }  \quad  i \in [1, k+1],  j  \in [1,h], t\in [1, (k+1) \ast h].
\end{equation}
Finally, the class whose similarity is highest is seen as the final prediction result, as shown in Eq~\eqref{WTA}.
\begin{equation}
\label{WTA}
pred = \mathit{argmax}\left ( S_{1}, S_{2}, ...,   S_{(k+1)\ast  h}\right ).
\end{equation}

% {\color{blue}
% With the assistance of Label Mapping, we hold a view that the similarity of the ground truth has an absolute advantage in all similarities so that it can be outputted as the final prediction.
% }

\subsubsection{Analysis of Label Mapping Algorithm}
\label{LM_analysis}
Label Mapping Algorithm can generate distinguishing label vectors as the learning targets of classes.
In order to accommodate continual arriving new classes, we need to investigate the capacity of Label Mapping Algorithm.
From the probability view,  we estimate the number of label vectors that can be generated by the Label Mapping Algorithm as following:
 \begin{equation}
N  = \frac{ log(1- e^{\frac{log(1-\tau)}{\gamma}} ) }{log(\mathrm{P(A)})} + 1
\label{eq:upper_bound_1}
 \end{equation}
where $\mathrm{P(A)} $ is:
\begin{equation}
\mathrm{P(A)}= \int_{-\infty }^{T}\frac{1}{\sqrt{2\pi(\frac{1}{d})^2}}e^{-\frac{S^2}{2(\frac{1}{d})^2}} dS
\label{eq:pa_1}
\end{equation}
where $T$, $\gamma$ and $d$  are consistent with the notations in Algorithm 1.
Eq~\eqref{eq:upper_bound_1} represents that we can take $N$ label vectors under the probability of $\tau$.  
If $d$ is large enough, or $T$ is not tight, the amount of label vectors is able to satisfy the requirements of general class incremental learning.
More details about the analysis of Label Mapping Algorithm can be found in the Appendix.

\subsection{Response Consolidation}

If the multi-head neural network with Label Mapping is trained incrementally without Rehearsal, it will still suffer from the catastrophic forgetting problem.
It is because that when we optimize the loss function of the new head, the output of old heads would definitely be affected by the shared parameters in the bottom layers.
Inspired by LwF \cite{li2017learning}, we propose Response Consolidation (RC) method to overcome catastrophic forgetting without the help of old classes.

We aim to make the outputs of the old heads remain stable during the new training process.
Therefore, we copy the entire model $\mathbf{M}$ as $\mathbf{\hat{M}}$ before training and freeze its weights.
Then we propagate the training data of new classes $\bm{X}$ through each head $\mathbf{\hat{M}_i}$ in $\mathbf{\hat{M}}$ and get the response vector $\bm{\hat{v}_i}$:
\begin{equation}
\label{eq2}
\bm{\hat{v}_i} \leftarrow \mathbf{\hat{M}_i}\left (\bm{x} \right ) \\
\end{equation}
Then we use $\bm{\hat{v}_i}$ as the \emph{memory targets} and maximize the cosine similarity between the $\bm{\hat{v}_i}$ and the output vector of the head $\mathbf{{M}_i}$ in $\mathbf{M}$, as Eq~\eqref{eq3}.
We call it as \textbf{Response Loss}.
In this way, we can force $\mathbf{{M}_i}$ to maintain the old behavior as $\mathbf{\hat{M}_i}$, so that we can keep the memory of the old classes.
\begin{equation}
\label{eq3}
\begin{aligned}
\bm{o_i} &\leftarrow \mathbf{M_i} \left ( \bm{x} \right ) \\
L_{i} &= \frac{ \bm{o_i} \cdot \bm{\hat{v}_i}}  { \left \| \bm{o_i} \right \| \   \left \| \bm{\hat{v}_i}\right \| } .
\end{aligned}
\end{equation}
Finally, we sum up the New Classes Loss $L_{k+1}$  in Eq~\eqref{equ:1} and the Response Loss $L_{i}(i=1,...,k)$ with a weight parameter $\lambda$ .
We define the loss function as follows.
\begin{equation}
\label{eq4}
Loss = - \left ( L_{k+1}+\lambda \sum_{i=1}^{k} \cdot L_{i} \right)
\end{equation}
\subsection{LMRC with Rehearsal Extension }
Although LMRC can work well without auxiliary data of the old classes, it can also be combined with various Rehearsal methods easily and flexibly. 
By means of Rehearsal, the model is able to review the information of the old classes to improve its accuracy.

Specifically, when the old classes are available, we can train each head using the data of all classes (assume $N$ classes in total), rather than merely those classes governed by itself. 
It can be formalized as Eq~\eqref{equ:all_class}.
We call it as \textbf{Review Loss}.
\begin{equation}
\label{equ:all_class}
\begin{aligned}
\bm{o_{i}^j} &\leftarrow  \mathbf{M_{i}}\left ( \bm{x}^j \right ) , \ j\in [1,N]  \\
L'_{i} &= \frac{  \bm{o_{i}^j} \cdot \bm{c}_{i}^{j} }{ \left \| \bm{o_{i}^j} \right \| \     \left \|  \bm{c}_{i}^{j}\right \| }, \  i \in [1,k]\\
\end{aligned}
\end{equation}
In this way, each head can get the information of other classes governed by other heads so that they can further learn the distinctiveness among all the classes by the $L'_{i} $. The final loss function is  Eq~\eqref{final_loss}, which is the combination of New Classes Loss, Response Loss, and Review Loss.
\begin{equation}
\label{final_loss}
Loss = - \left ( L_{k+1}+ \sum_{i=1}^{k} \cdot L'_{i} +\lambda \sum_{i=1}^{k} \cdot L_{i} \right)
\end{equation}
The testing phase is still the same as Fig.~\ref{LMRC} (Right). 
Note that LMRC utilizes Label Mapping instead of one-hot coding. 
It means that no matter how many classes there are, the dimensionality of label vectors are fixed. 
Even if we increase the number of classes to be learned on each head,\textbf{ the network structure does not need to be changed at all}.
Therefore, except for the input data, the training and testing processes are exactly the same as those of standard LMRC.

\section{Experiment}

\label{experiment}
In order to evaluate the effectiveness of LMRC, we conduct a series of experiments on two benchmark datasets, including CIFAR-100 \cite{krizhevsky2009learning} and ImageNet-200 \cite{yao2015tiny}. 
 We compare the accuracy of LMRC with that of other related methods. 
After that, we validate the advantages of Label Mapping and Response Consolidation, respectively. 
The experimental source code can be referred in \href{https://github.com/personal-paper-code/LMRC}{https://github.com/personal-paper-code/LMRC}.

\subsection{Setup and Implementation}
We conduct experiments on two benchmark datasets.
Each dataset is divided into several parts by classes. 
Each part, called as a \emph{Class Batch}, contains the data of several classes that do not overlap with other parts. 
In CIFAR-100/ImageNet-200, we divide 100/200 classes into 10 class batches, respectively. 
The training sets of these class batches are fed into the model sequentially for training.
Note that when the model is being trained on the current class batch, the previous class batches cannot be accessed anymore. 
After that, we predict the testing sets of all classes that have been trained.
Finally, we calculate the \emph{average incremental accuracy}~\cite{castro2018end-to-end} of all class batches. 
All the reported results are the average accuracies of 5 executions of experiments.
Particularly, we report the top-5 accuracy on Imagenet-200 dataset.
For the fairness of comparison, we fix the architecture of the basic neural network and related hyper-parameters strictly. 
More details about the experiment are introduced in the Appendix.

\subsection{Effectiveness of LMRC}
\label{SEC_RQ1}

\begin{figure}[h!]
\vspace{-10pt}
\centering
\subfigure[CIFAR-100]
{
\includegraphics[height=0.25\linewidth,width=0.4\linewidth]{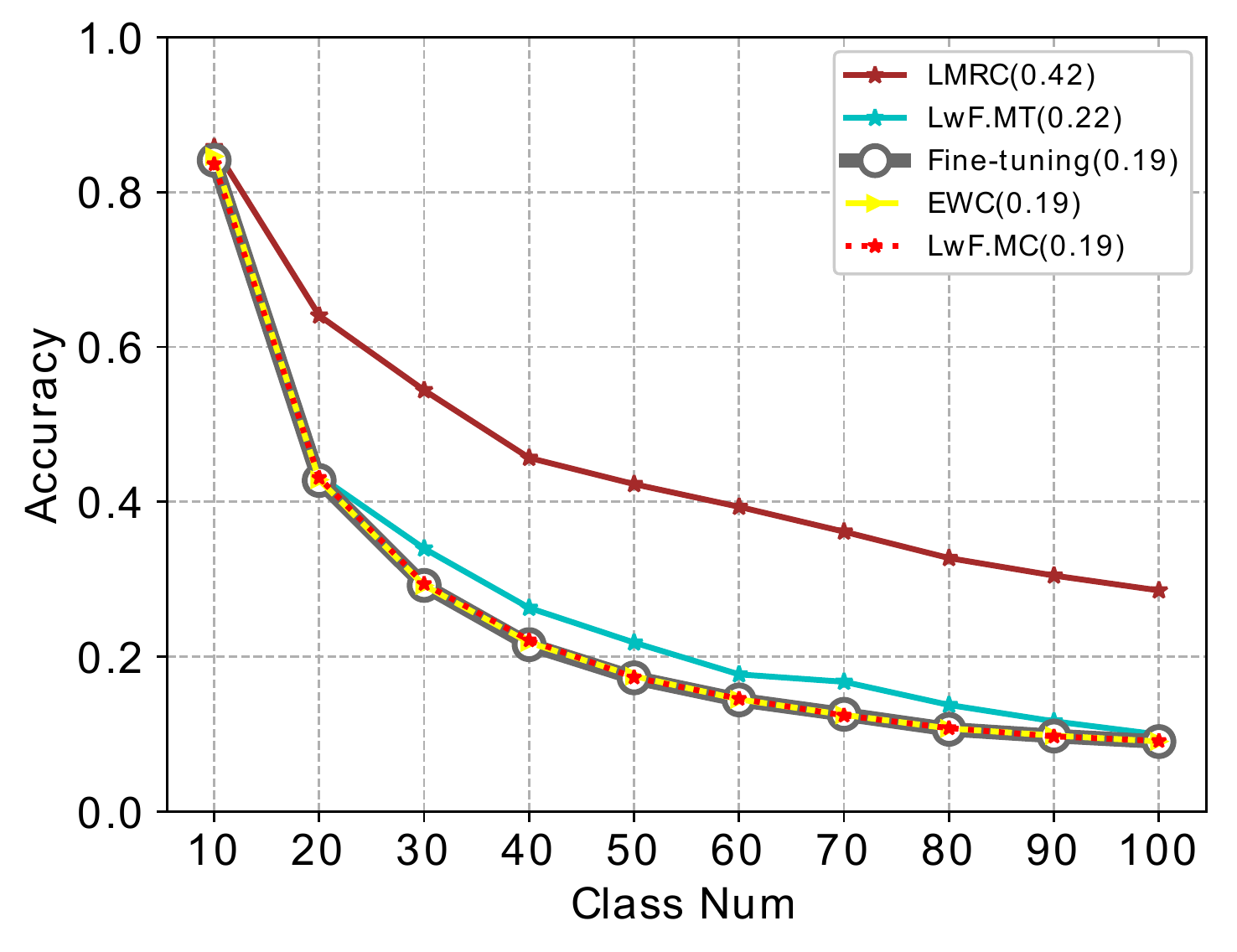}
}
\subfigure[Imagenet-200]
{
\includegraphics[height=0.25\linewidth,width=0.4\linewidth]{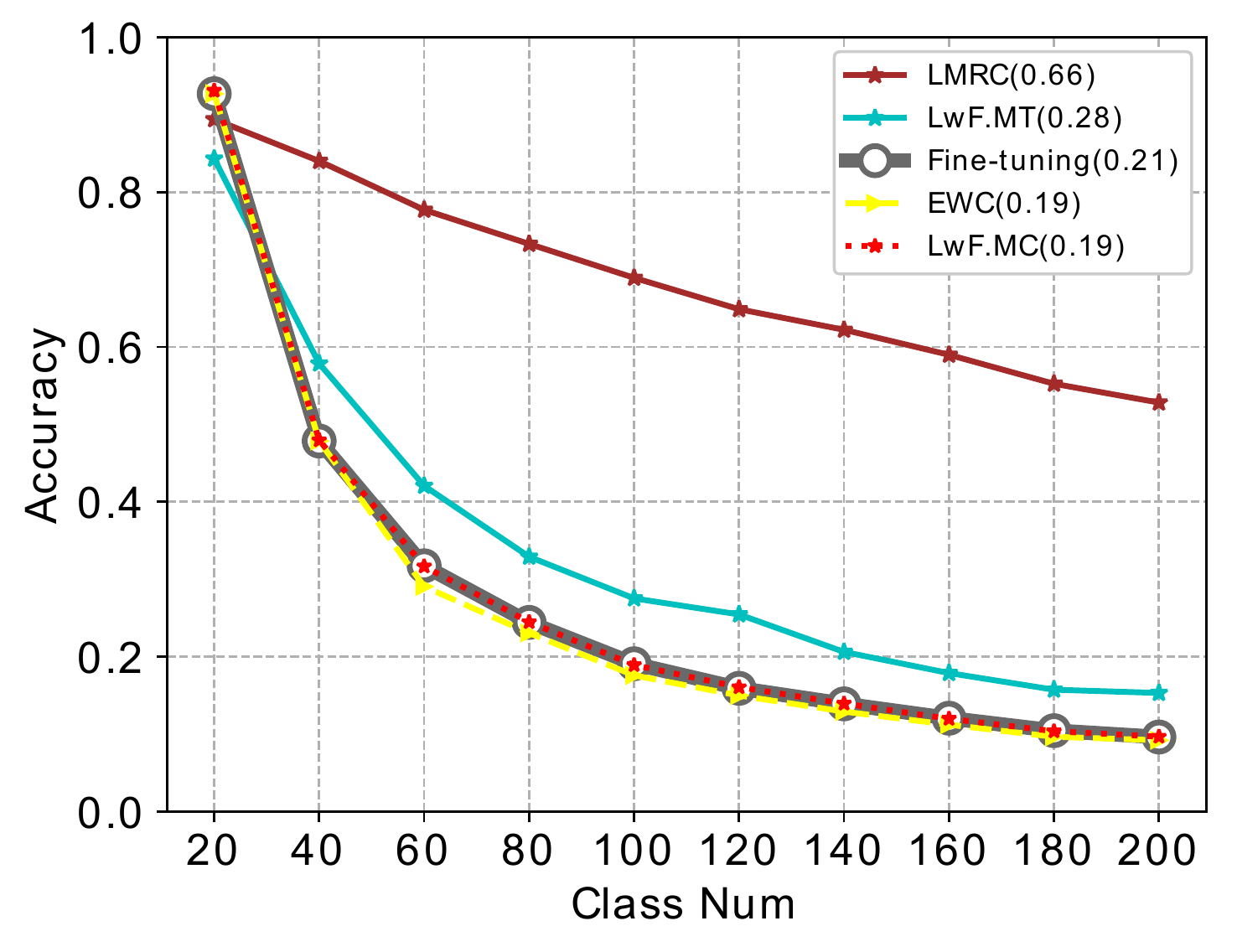}
}
\caption{The accuracies of LMRC and related methods on all class batches.}
\label{RQ1}
\vspace{-15pt}
\end{figure}

We have implemented four comparative methods, including Fine-tuning, EWC ~\cite{kirkpatrick2017overcoming}, LwF.MC~\cite{castro2018end-to-end}  and LwF.MT~\cite{li2017learning} to validate the effectiveness of LMRC.
Fine-tuning means that when new classes arrive, we only expand the dimension of softmax layer and one-hot codes for training the new classes, but take no measures to overcome the softmax suppression and catastrophic forgetting problems.
In EWC/LwF.MC, we not only expand the dimension but also use the EWC/LwF algorithm  to handle the catastrophic forgetting problem.
LwF.MT refers to the multi-head implementation of LwF.
It is similar to LMRC but uses the traditional softmax layer in all heads instead of Label Mapping. 
The accuracies of LMRC and the related methods on all class batches are shown in Fig.~\ref{RQ1}.
The average incremental accuracy is shown in the legend. 

From Fig.~\ref{RQ1},  we can see that LwF.MC and EWC perform as poorly as Fine-tuning, which only remember the information of the current class batch.
The results show that even if those methods could overcome catastrophic forgetting, the softmax suppression problem in class incremental learning remains an obstacle for them.
In addition, we  noted that LwF.MT performs better than the above three methods.
We conclude that the multi-head architecture is effective for class incremental learning.
LMRC achieves the highest accuracy in all class batches of all datasets.
For example, in ImageNet-200 dataset, LMRC can achieve a much higher average incremental accuracy of $66.4\%$, which exceeds the other methods by about $40\%$.
Most importantly, these remarkable results of LMRC are obtained without any assistance of Rehearsal.

% \begin{figure}[htbp]
% \centering
% \subfloat[CIFAR-100]{
% \begin{minipage}[t]{0.48\textwidth}
% \centering
% \includegraphics[height=0.65\textwidth,width=1\textwidth]{image/RQ1_CIFAR100.eps}
% \label{LMRC}
% \end{minipage}
% }
% \subfloat[ImageNet-200]{
% \begin{minipage}[t]{0.48\textwidth}
% \centering
% \includegraphics[height=0.65\textwidth,width=1\textwidth]{image/RQ1_Imagenet200.eps}
% \label{predict}
% \end{minipage}
% }
% \caption{The accuracies of LMRC and related methods on every class batch.}
% \end{figure}

\subsection{Effectiveness of LMRC after extending with Rehearsal}
\label{SEC_RQ2}

If allowed, we can extend LMRC with Rehearsal, in which our model is allowed to access the previous data of old classes.
In this experiment, we use a data pool as the Rehearsal component.
Considering the fairness and universality, we randomly select several samples from each old class and add them into a data pool, without any elaborate sampling strategy.
We combine the data pool with all comparison methods in Sec.~\ref{SEC_RQ1} except LwF.MT,  which is not suitable for using old data. 
The experimental results are shown in Table. \ref{LMRC_datapool}.

Comparing with Fig.~\ref{RQ1}, it is clear that Rehearsal improves the accuracy of all models.
However, LMRC can still achieve the best accuracy in all datasets.
We have also noticed that as we decrease the size of data pool (from 100 to 20), LMRC achieves a much larger marginal advantage compared with the related methods.
It means that LMRC can perform better with much fewer samples of old classes.

\begin{table}[h]
\vspace{-10pt}
\footnotesize
\centering
\setlength{\abovecaptionskip}{0pt}
 \setlength{\belowcaptionskip}{10pt}
\caption{Average incremental accuracy of class incremental models with Rehearsal extension}
%\vspace{5}
\begin{tabular}{cc|ccc|ccc}
    \toprule
    \multicolumn{1}{c|}{\multirow{2}[3]{*}{Model}} & Dataset  & \multicolumn{3}{c|}{CIFAR-100} & \multicolumn{3}{c}{ImageNet-200} \\
\cmidrule{2-8}    \multicolumn{1}{c|}{} & Sample Num  & 20      & 50      & 100     & 20      & 50      & 100 \\
\midrule
    \multicolumn{2}{c|}{Fine-tuning}  & 0.392   & 0.513   & 0.606   & 0.563   & 0.671   & 0.721  \\
    \multicolumn{2}{c|}{EWC}    & 0.408   & 0.522   & 0.602   & 0.557   & 0.673   & 0.728  \\
    \multicolumn{2}{c|}{LwF.MC} & 0.402   & 0.516   & 0.607   & 0.564   & 0.665   & 0.731  \\
    \multicolumn{2}{c|}{LMRC}    & \textbf{0.497}  & \textbf{0.542}   &\textbf{ 0.622}   & \textbf{0.662}   &\textbf{ 0.691 }  & \textbf{0.738 } \\
     %\multicolumn{2}{c|}{iCarl} & 0.950   & 0.963   & 0.967   & 0.564  & 0.692   & 0.740   & 0.493   & 0.575   & 0.621   & 0.650   & 0.701   & 0.736 \\
    \bottomrule
    \end{tabular}%
\label{LMRC_datapool}
\vspace{-10pt}
\end{table}
% & \multicolumn{3}{c|}{MNIST}  & \multicolumn{3}{c|}{CIFAR-10}
%  & 20      & 50      & 100     & 20      & 50      & 100   
%  & 0.875   & 0.939   & 0.966   & 0.429   & 0.623   & 0.670  
%  & 0.864   & 0.940   & 0.964   & 0.480   & 0.592   & 0.672
%   & 0.849   & 0.935   & 0.968   & 0.479   & 0.608   & 0.681  
%   & \textbf{0.972}   & \textbf{0.973 }  & \textbf{0.973}   & \textbf{0.542}   & \textbf{0.645 }  & \textbf{0.720}
\subsection{Effectiveness of Label Mapping}
\label{SEC_RQ3}
\subsubsection{Compare with LwF.MT}
LwF.MT and LMRC both utilize multi-head network.
One of the differences between LwF.MT and LMRC is that the latter uses Label Mapping but  LwF.MT uses one-hot coding.
By comparing LMRC with  LwF.MT, we can evaluate the role of Label Mapping in class incremental learning.
From Fig. \ref{RQ1}, it is can be seen that LMRC performs much better than  LwF.MT in all class batches.
We also noticed that  LwF.MT performs better in the early stage than the later.
It is because that the probabilities come from different heads and are not easily confused under a small number of classes.
However, the confusion effect would become more significant as the number of classes increases.
In contrast, LMRC can maintain a much higher accuracy because we use Label Mapping to generate label vectors, which can achieve high distinctiveness among the heads.
\subsubsection{Class Capacity}
\label{Class_Capacity}
\begin{figure}[h!]
\centering
\subfigure[Class Capacity]
{\includegraphics[height=0.2\linewidth,width=0.3\linewidth]{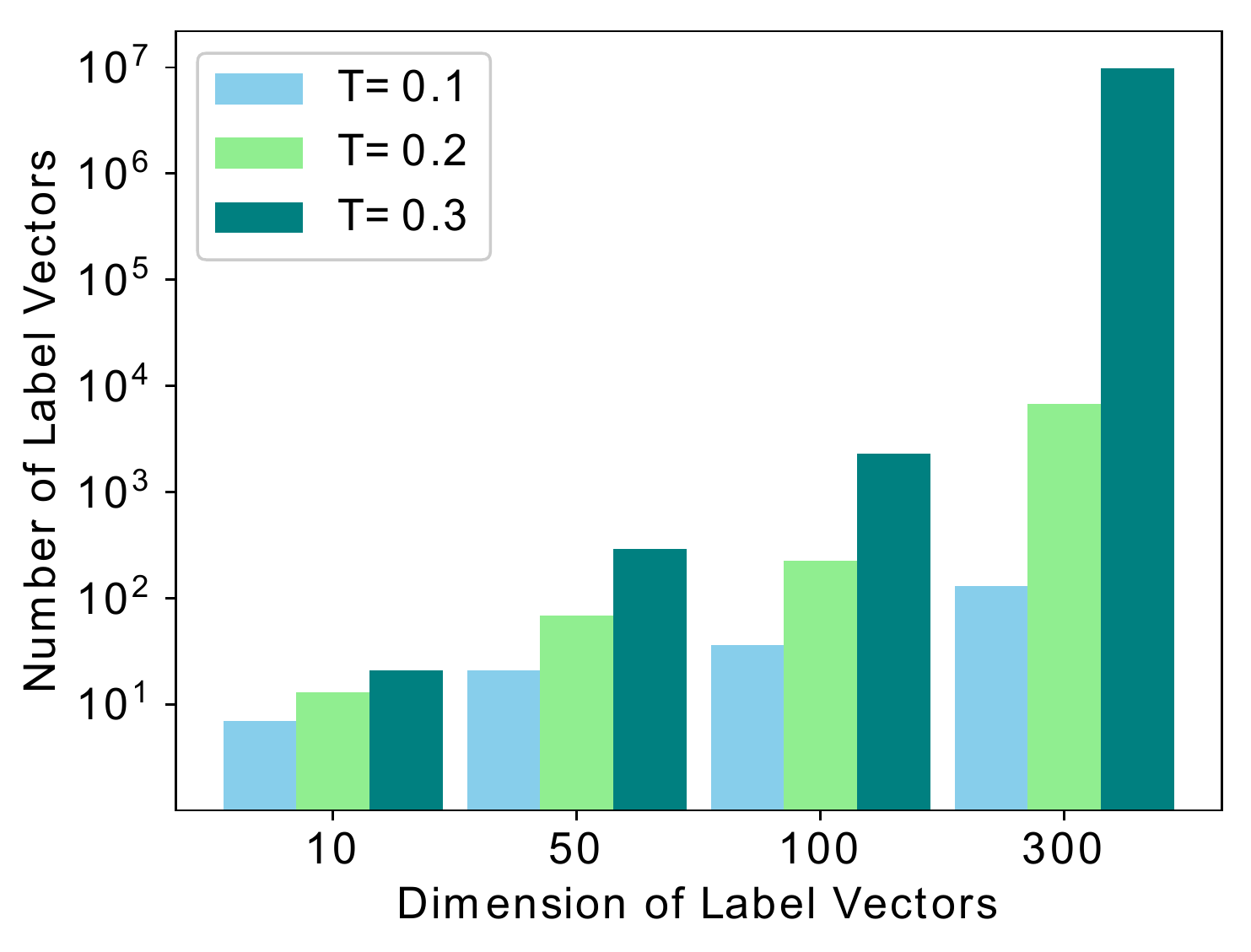}}
%\label{capacity}
\subfigure[CIFAR-100]
{
\includegraphics[height=0.2\linewidth,width=0.3\linewidth]{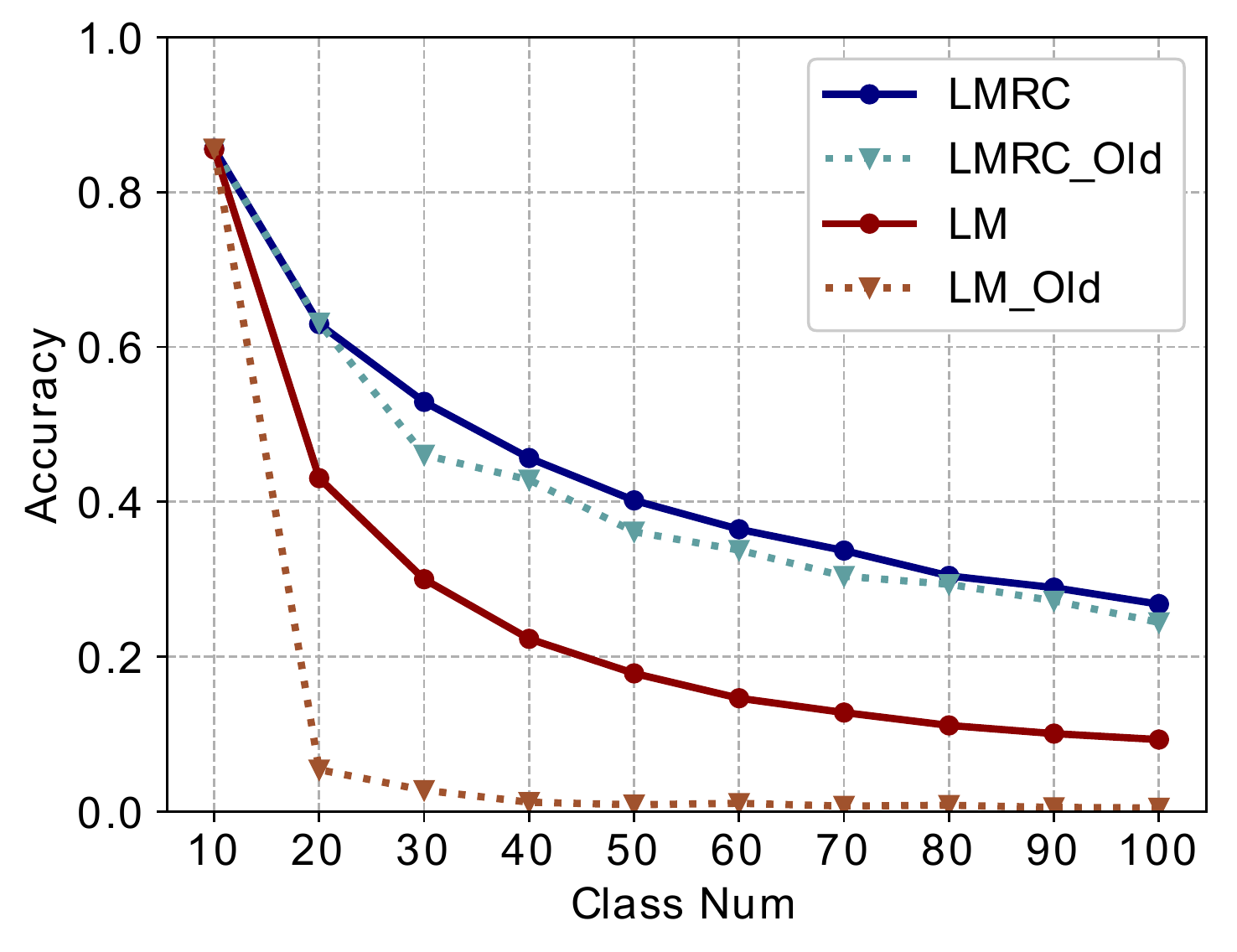}
%\label{RC_CIFAR100}
}
\subfigure[Imagenet-200]
{
\includegraphics[height=0.2\linewidth,width=0.3\linewidth]{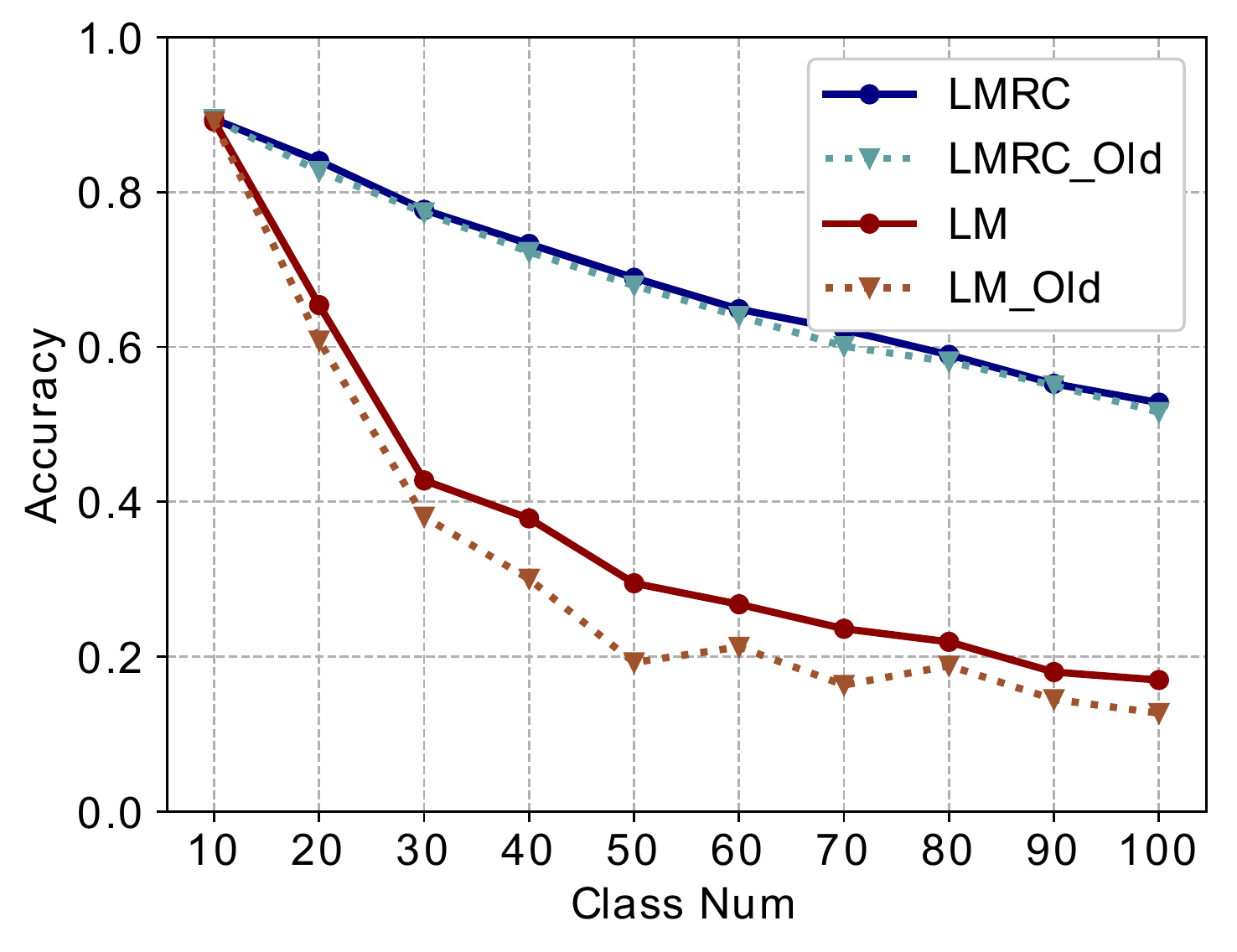}
%\label{RC_Imagenet200}
}
\caption{\textbf{(a)}: Class capacity of Label Mapping  under different dimension $d$ and threshold $T$. \textbf{(b) \& (c)}: Classification accuracy of LMRC and LM (CIFAR-100 \& ImageNet-200).}
\label{RQ2}
\vspace{-15pt}
\end{figure}

Theoretically, a class incremental learning model should have no limitation on the class amount.
Thus, we conduct an experiment to explore how many label vectors can be generated by the Label Mapping algorithm.
We specify the maximum random sampling times $\gamma = 10000$ and count how many label vectors can be found out.
The result is shown in Fig.~\ref{RQ2}~(a).
It is evident that when the dimensionality $d$ of vector space increases, or we set a looser similarity threshold $T$,  it is much easier to obtain more label vectors.
We can easily generate $10^3\sim 10^7$ label vectors by our algorithm, which can meet the requirement of the  class incremental learning task.
We also show the theoretical result of Label Mapping in the Appendix.
It is basically consistent with our empirical experiment result.
% \begin{figure}[htbp]
% \centering
% \subfloat[]{
% \begin{minipage}[t]{0.3\linewidth}
% \centering
% \includegraphics[height=0.77\textwidth,width=\textwidth]{image/LM_contain.eps}
% \end{minipage}
% }
% \subfloat[]{
% \label{capacity}
% \begin{minipage}[t]{0.3\linewidth}
% \centering
% \includegraphics[height=0.7\textwidth,width=\textwidth]{image/RC_CIFAR100.eps}
% \label{RC_CIFAR100}
% \end{minipage}
% }
% \subfloat[]{
% \begin{minipage}[t]{0.3\linewidth}
% \centering
% \includegraphics[height=0.7\textwidth,width=\textwidth]{image/RC_Imagenet200.eps}
% \label{RC_Imagenet200}
% \end{minipage}
% }
%  \caption{\textbf{Left}: Class capacity of Label Mapping  under different dimension $d$ and threshold $T$. \textbf{Middle}: Classification accuracy of LMRC and LM (CIFAR-100). \textbf{Right}: Classification accuracy of LMRC and LM (ImageNet-200).} 
% \end{figure}

\subsection{Effectiveness of Response Consolidation}
\label{SEC_RQ4}
In order to evaluate the effectiveness of Response Consolidation, we compare the accuracy of LMRC and our proposed model without RC (denoted as LM). 
The result is illustrated in Fig.~\ref{RQ2} (a) and (b).
It shows that the accuracy of LMRC is much higher than that of LM. Furthermore, we also conclude that the main cause of this result lies in the accuracy decline on old classes.
We also illustrate the  classification accuracy of old classes (dotted line) in Fig.~\ref{RQ2} (a) and (b).
It is obvious that the LM cannot memorize the information of the old classes, and therefore its total accuracy is much lower.
On the contrary, LMRC is able to keep the information of  the old classes very well.
Thus it confirms the validity of Response Consolidation.

\section{Conclusion}

In this paper,  we propose a new class incremental learning model, named Label Mapping with Response Consolidation (LMRC).
It can work well without accessing the data of old classes. 
We propose the Label Mapping algorithm  for mitigating the softmax suppression problem and  propose the Response Consolidation method to handle the catastrophic forgetting problem. 
In our experiment, LMRC achieves remarkable results compared with  the related methods.

\bibliographystyle{plain}
\bibliography{ref}

\begin{thebibliography}{10}

\bibitem{castro2018end-to-end}
Francisco~M Castro, Manuel~J Marinjimenez, Nicolas Guil, Cordelia Schmid, and
  Karteek Alahari.
\newblock End-to-end incremental learning.
\newblock {\em european conference on computer vision}, pages 241--257, 2018.

\bibitem{degroot2012probability}
Morris~H DeGroot and Mark~J Schervish.
\newblock {\em Probability and statistics}.
\newblock Pearson Education, 2012.

\bibitem{he2016deep}
Kaiming He, Xiangyu Zhang, Shaoqing Ren, and Jian Sun.
\newblock Deep residual learning for image recognition.
\newblock In {\em Proceedings of the IEEE conference on computer vision and
  pattern recognition}, pages 770--778, 2016.

\bibitem{kiefer1952stochastic}
J~Kiefer and J~Wolfowitz.
\newblock Stochastic estimation of the maximum of a regression function.
\newblock {\em Annals of Mathematical Statistics}, 23(3):462--466, 1952.

\bibitem{kirkpatrick2017overcoming}
James Kirkpatrick, Razvan Pascanu, Neil Rabinowitz, Joel Veness, Guillaume
  Desjardins, Andrei~A Rusu, Kieran Milan, John Quan, Tiago Ramalho, Agnieszka
  Grabska-Barwinska, et~al.
\newblock Overcoming catastrophic forgetting in neural networks.
\newblock {\em Proceedings of the national academy of sciences}, page
  201611835, 2017.

\bibitem{krizhevsky2009learning}
Alex Krizhevsky and Geoffrey Hinton.
\newblock Learning multiple layers of features from tiny images.
\newblock Technical report, Citeseer, 2009.

\bibitem{krizhevsky2012imagenet}
Alex Krizhevsky, Ilya Sutskever, and Geoffrey~E Hinton.
\newblock Imagenet classification with deep convolutional neural networks.
\newblock In {\em Advances in neural information processing systems}, pages
  1097--1105, 2012.

\bibitem{li2017learning}
Zhizhong Li and Derek Hoiem.
\newblock Learning without forgetting.
\newblock {\em european conference on computer vision}, pages 614--629, 2017.

\bibitem{mccloskey1989catastrophic}
Michael McCloskey and Neal~J Cohen.
\newblock Catastrophic interference in connectionist networks: The sequential
  learning problem.
\newblock In {\em Psychology of learning and motivation}, volume~24, pages
  109--165. Elsevier, 1989.

\bibitem{muller1959note}
Mervin~E Muller.
\newblock A note on a method for generating points uniformly on n-dimensional
  spheres.
\newblock {\em Communications of the ACM}, 2(4):19--20, 1959.

\bibitem{robins1998catastrophic}
Anthony Robins and SIMON McCALLUM.
\newblock Catastrophic forgetting and the pseudorehearsal solution in
  hopfield-type networks.
\newblock {\em Connection Science}, 10(2):121--135, 1998.

\bibitem{shin2017continual}
Hanul Shin, Jung~Kwon Lee, Jaehong Kim, and Jiwon Kim.
\newblock Continual learning with deep generative replay.
\newblock {\em neural information processing systems}, pages 2990--2999, 2017.

\bibitem{van2018generative}
Gido~M van~de Ven and Andreas~S Tolias.
\newblock Generative replay with feedback connections as a general strategy for
  continual learning.
\newblock {\em arXiv preprint arXiv:1809.10635}, 2018.

\bibitem{wen2016discriminative}
Yandong Wen, Kaipeng Zhang, Zhifeng Li, and Yu~Qiao.
\newblock A discriminative feature learning approach for deep face recognition.
\newblock In {\em European conference on computer vision}, pages 499--515.
  Springer, 2016.

\bibitem{85977}
whuber (https://stats.stackexchange.com/users/919/whuber).
\newblock Distribution of scalar products of two random unit vectors in $d$
  dimensions.
\newblock Cross Validated.
\newblock URL:https://stats.stackexchange.com/q/85977 (version: 2017-04-13).

\bibitem{xiao2014error}
Tianjun Xiao, Jiaxing Zhang, Kuiyuan Yang, Yuxin Peng, and Zheng Zhang.
\newblock Error-driven incremental learning in deep convolutional neural
  network for large-scale image classification.
\newblock In {\em Proceedings of the 22nd ACM international conference on
  Multimedia}, pages 177--186. ACM, 2014.

\bibitem{yao2015tiny}
Leon Yao and John Miller.
\newblock Tiny imagenet classification with convolutional neural networks.
\newblock {\em CS 231N}, 2015.

\bibitem{zenke2017continual}
Friedemann Zenke, Ben Poole, and Surya Ganguli.
\newblock Continual learning through synaptic intelligence.
\newblock {\em international conference on machine learning}, 8:3987--3995,
  2017.

\bibitem{zhou2002hybrid}
Zhi-Hua Zhou and Zhao-Qian Chen.
\newblock Hybrid decision tree.
\newblock {\em Knowledge-based systems}, 15(8):515--528, 2002.

\end{thebibliography}

\newpage
\section{Appendix}

\subsection{Experiment Implementation Details}
For the fairness of comparison, we fix the  neural network architecture and related hyper-parameters strictly. 
We build a ResNet-18 \cite{he2016deep} as the basic CNN structure.
The networks are optimized by the Stochastic Gradient Descent (SGD) algorithm \cite{kiefer1952stochastic} with a batch size of 128. 
In  CIFAR-100 experiment, the epoch is set to 70. 
The learning rate begins with 0.05 and is halved at the $50$th and $60$th epoch.
In  ImageNet-200 experiment, the epoch is set to 80. 
The learning rate begins with 0.1 and is halved at the $50$th, $65$th and $75$th epoch.
The similarity threshold $T$ in Label Mapping is set to $0.15 \sim  0.20$ and $\lambda$ in Eq~\eqref{eq4} is set to $1 \sim 5$. 
The dimension of label vectors $d$ is fixed as 100.

\subsection{Analysis of Label Mapping Algorithm}

Label Mapping Algorithm is able to generate distinguishing label vectors as the learning targets of different classes. 
Intuitively, the larger the dimension $d$, the more label vectors can be generated by the Label Mapping. 
Meanwhile, a larger threshold parameter $T$ means a much looser restriction on orthogonality, hence more label vectors can also be generated.
In this section, we conduct a simple theoretical analysis  to confirm this intuition.
First of all, we give two lemmas for auxiliary of the analysis. 

\textbf{Lemma1}:
Let random variables $X_1, X_2,..., X_n \sim \mathcal{N}(0,1)$ and be independent. 
The vector $\bm{X} = \left(\frac{X_1}{Z}, \frac{X_2}{Z}, \cdots, \frac{X_n}{Z}\right)$
is \textbf{uniformly} distributed on the surface of the hyper-sphere $S_{n-1}$, where $Z = \sqrt{X_1^2 + \cdots + X_n^2}$ is a normalization factor.
The proof of Lemma 1 can be referred in ~\cite{muller1959note}.
Actually, we  randomly sample the vectors in Line \ref{marker} of Algorithm \ref{alg1}  according to this Lemma.

\textbf{Lemma2}:
Let $t$ denotes the inner product of two \textbf{uniformly} random unit vectors:  $\bm{x} \in \mathbb{R}^d$ and $\bm{y} \in \mathbb{R}^d$ .
We can conclude that $u= \frac{t+1}{2}$  follows a $Beta(\frac{d-1}{2},\frac{d-1}{2})$ distribution and the standardized distribution of $t$ approaches normality at a rate of $O(\frac{1}{d})$.
The proof of Lemma 2 can be referred in ~\cite{85977}

According to the Lemma 2, we can conclude that  $t$ approximately follows a Gaussian distribution with $\mu=0$ and $\sigma=\frac{1}{d}$, i.e. 
\begin{equation}
t  \sim \mathcal {N}(0,\frac{1}{d})
\end{equation}

We set the Event $\mathrm{A}$ as that the cosine similarity of  two uniformly random unit vectors $\bm{x} \in \mathbb{R}^d$ and $\bm{y} \in \mathbb{R}^d$  is less than threshold $T$.
The probability of $\mathrm{A}$ is as following:
\begin{equation}
\mathrm{P(A)} = \left \{ S =  \frac{\bm{x}\cdot \bm{y}}{\left \| \bm{x} \right \| \left \| \bm{y} \right \|}\leqslant T \right \}
\end{equation}
Note that $\bm{x}$ and $\bm{y}$ are unit vectors so we can get:

\begin{equation}
\mathrm{P(A)} = \left \{ S =   \bm{x} \cdot \bm{y} \leqslant T \right \}
\end{equation}

According to the Lemma 2, we have known that $S\sim \mathcal {N}(0,\frac{1}{d})$.
Therefore, we can calculate the $\mathrm{P(A)}$ as following:

\begin{equation}
\mathrm{P(A)} = \int_{-\infty }^{T}\frac{1}{\sqrt{2\pi(\frac{1}{d})^2}}e^{-\frac{S^2}{2(\frac{1}{d})^2}} dS
\label{eq:pa}
\end{equation}
In the $n$-th step of Algorithm~\ref{alg1}, we try to get the $n$-th label vectors when we have already obtained  $n-1$ label vectors in the candidate set.
Once we get $\mathrm{P(A)}$, we can calculate the success probability of \textbf{$n$-step}.
We denote the Event B that means we successfully get the $n$-th label vector at once.
To make Event B happen, we only need to guarantee those cosine similarities between the new randomly sampled vector and the other  $n-1$ label vectors in candidate set are less than $T$.
It is worth noting that the existing $n-1$ label vectors are approximately orthogonal.
Thus, we assume that the comparison between each candidate label vector and the randomly sampled vector is approximately independent.
It can be formalized as following, where $\bm{v}_s$ is the new randomly sampled vector and $\bm{c}_i$ denotes the label vector in the candidate set.
 \begin{equation}
  \begin{aligned}
%  \mathrm{P(B)} = \prod_{1}^{n-1}\mathrm{P(A)} = \mathrm{P(A)}^{(n-1)}
\mathrm{P(B)} &= \mathrm{P}\left \{ \bm{c}_1 \cdot \bm{v}_s \leqslant T, \bm{c}_2 \cdot \bm{v}_s \leqslant T,...,\bm{c}_{n-1} \cdot \bm{v}_s \leqslant T \right \}\\
&\approx \mathrm{P}\left \{ \bm{c}_1 \cdot \bm{v}_s \leqslant T\right \}\cdot \mathrm{P}\left \{ \bm{c}_2 \cdot \bm{v}_s \leqslant T\right \}\cdot ...\cdot \mathrm{P}\left \{ \bm{c}_{n-1}  \cdot \bm{v}_s \leqslant T\right \} \\
&=\prod_{1}^{n-1}\mathrm{P(A)} = \mathrm{P(A)}^{(n-1)}
 \label{eq:pb}
 \end{aligned}
 \end{equation}
 From the Eq~\eqref{eq:pb}, we can see that the probability of Event B decreases exponentially with the increase of $n$.
 It indicates that it becomes more and more difficult to get the $n$-th label vectors as $n$ increasing.
 Therefore, in the algorithm, we try to sample for $\gamma $ times  in order to get the $n$-th label vector. 
 The success probability of each time is $\mathrm{P(B)}$ .
We set  a random variable $K$,  which represents that we get the $n$-th label vector in the $k$-th trial.
Obviously, $K$ follows the Geometric Distribution~\cite{degroot2012probability}.
 Its Cumulative Distribution Function (CDF) is as following:
 \begin{equation}
 \mathrm{P}(K\leqslant k) = 1-(1-\mathrm{P(B)})^{k}
 \label{eq:cdf}
 \end{equation}
Let $k = \gamma$ , we can get the probability of getting $n$-th label vector by means of Label Mapping Algorithm.
 \begin{equation}
\mathrm{P} (K\leqslant \gamma) = 1-(1-\mathrm{P(B)})^{\gamma}
 \label{eq:final_p}
 \end{equation}
 From the Eq~\eqref{eq:pa}, Eq~\eqref{eq:pb} and Eq~\eqref{eq:final_p}, we can see that as the $T$ or $d$ increasing, $\mathrm{P(A)}$, $\mathrm{P(B)}$  and $\mathrm{P} (K\leqslant \gamma)$   would become higher accordingly. 
 
We expect that  we can get the $n$-th label vector under a high probability, i.e.  $\mathrm{P} (K\leqslant \gamma)  \geqslant \tau $, where $\tau$ is a probability threshold. 
Therefore, we can estimate the number of label vectors $N$ as following:
 \begin{equation}
 \begin{aligned}
 & \mathrm{P} (K\leqslant \gamma)   = 1 - (1- \mathrm{P(B)})^{\gamma } \geqslant \tau \\
 \Rightarrow  &(1-\mathrm{P(B)})^{\gamma} \leqslant 1-\tau \\
\Rightarrow &\gamma log (1-\mathrm{P(B)}) \leqslant log(1-\tau)\\
 \Rightarrow & \mathrm{P(B)} \geqslant   1- e^{\frac{log(1-\tau)}{\gamma}}  \\
\Rightarrow  & \mathrm{P(A)}^{(n-1)} \geqslant 1- e^{\frac{log(1-\tau)}{\gamma}}  \\
\Rightarrow &(n-1) \cdot log(\mathrm{P(A)}) \geqslant log(1- e^{\frac{log(1-\tau)}{\gamma}} ) \\
\Rightarrow &n  \leqslant \frac{ log(1- e^{\frac{log(1-\tau)}{\gamma}} ) }{log(\mathrm{P(A)})} + 1  \qquad  \because (log(\mathrm{P(A)} < 0)
 \end{aligned}
\end{equation}

Once $T$ and $d$ are specified in advance, $\mathrm{P(A)}$ is also fixed according to Eq~\eqref{eq:pa}.
Thus, we can get the theoretical number of label vectors $N$ is:
 \begin{equation}
N  = \frac{ log(1- e^{\frac{log(1-\tau)}{\gamma}} ) }{log(\mathrm{P(A)})} + 1
\label{eq:upper_bound}
 \end{equation}
It means that we can take $N$ label vectors under the probability of $\tau$. 

\begin{figure}[H]
  \centering
  \includegraphics[height=0.35\textwidth,width=.55\textwidth]{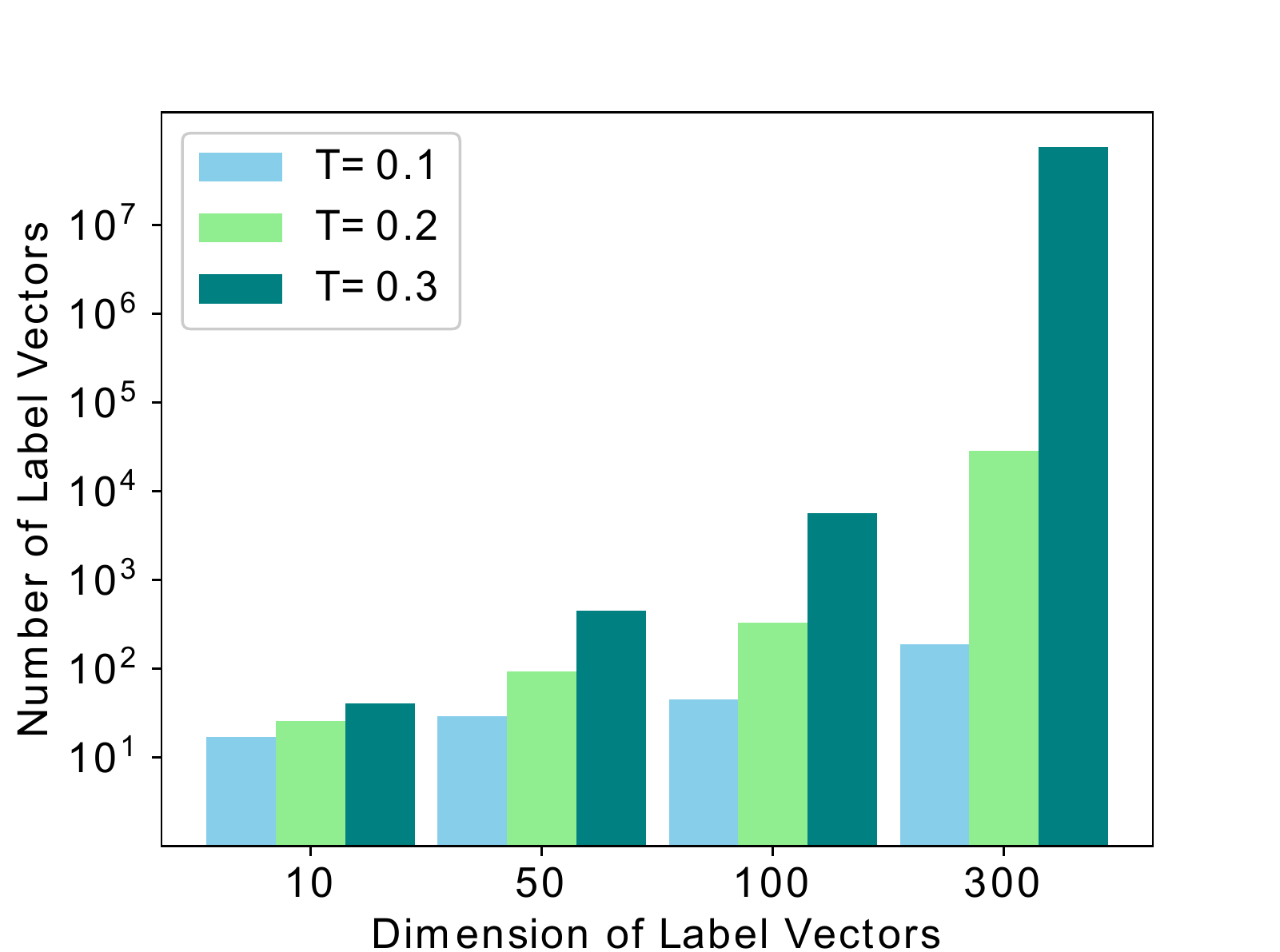}
  \caption{Theoretical capacity of Label Mapping  under different dimension $d$ and threshold $T$
  }
  \label{capacity_upper_bound}
\end{figure}

We set different $T$ and $d$ to observe the changes of $N$.
In Sec.~\ref{Class_Capacity}, we specify the maximum random sampling times $\gamma$ to 10000 and count how many label vectors can be found out.
In this section, we also set $\gamma = 10000$ and  $\tau = 0.99$. 
We calculate  $N$ according to Eq~\eqref{eq:upper_bound}.
The experimental result is given in Fig.~\ref{capacity_upper_bound}.
Obviously, it is consistent with the experimental result in Sec.~\ref{Class_Capacity}.

\end{document}